\documentclass[letterpaper, 10 pt, conference]{ieeeconf}  
\usepackage{tabulary,graphicx,times,caption,fancyhdr,amsfonts,amssymb,amsbsy,latexsym,amsmath}
\usepackage[utf8]{inputenc}
\usepackage{url,multirow,morefloats,floatflt,cancel,tfrupee,textcomp,colortbl,xcolor,pifont}
\usepackage{siunitx}
\usepackage[nointegrals]{wasysym}
\usepackage{hyperref}
\usepackage[ruled]{algorithm2e}
\usepackage{booktabs}
\usepackage{subfigure}
\usepackage{array}
\usepackage{bbm}
\usepackage{makecell}
\usepackage{textcomp}
\usepackage{url}
\usepackage{verbatim}
\usepackage{dutchcal}
\usepackage{extarrows}
\usepackage{multicol}
\usepackage{cite}
\usepackage{hyperref} 

\IEEEoverridecommandlockouts                              

\title{\LARGE \bf
MTD-GPT: A Multi-Task Decision-Making GPT Model for Autonomous Driving at Unsignalized Intersections
}

\author{Jiaqi Liu, Peng Hang,~\IEEEmembership{Member,~IEEE,} Xiao Qi , Jianqiang Wang, and Jian Sun 
\thanks{This work was supported in part by the Natural Science Foundation of China (52232015 and 52125208), the Young Elite Scientists Sponsorship Program by CAST (2022QNRC001) and the Fundamental Research Fund for the Central Universities.}
\thanks{Jiaqi Liu, Peng Hang, Xiao Qi and Jian Sun are with the Department of
Traffic Engineering and Key Laboratory of Road and Traffic Engineering,
Ministry of Education, Tongji University, Shanghai 201804, China. (e-mail: \{liujiaqi13, hangpeng, qixiao, sunjian\}@tongji.edu.cn)}
\thanks{Jianqiang Wang is with State Key Laboratory of Automotive Safety and Energy
School of Vehicle and Mobility, Tsinghua University, Beijing
100084, China.(e-mail: wjqlws@tsinghua.edu.cn)}
\thanks{Corresponding author: Peng Hang}
}

\begin{document}

\maketitle
\thispagestyle{empty}
\pagestyle{empty}

\begin{abstract}

Autonomous driving technology is poised to transform transportation systems. However, achieving safe and accurate multi-task decision-making in complex scenarios, such as unsignalized intersections, remains a challenge for autonomous vehicles. This paper presents a novel approach to this issue with the development of a Multi-Task Decision-Making Generative Pre-trained Transformer (MTD-GPT) model. Leveraging the inherent strengths of reinforcement learning (RL) and the sophisticated sequence modeling capabilities of the Generative Pre-trained Transformer (GPT), the MTD-GPT model is designed to simultaneously manage multiple driving tasks, such as left turns, straight-ahead driving, and right turns at unsignalized intersections.
We initially train a single-task RL expert model, sample expert data in the environment, and subsequently utilize a mixed multi-task dataset for offline GPT training. This approach abstracts the multi-task decision-making problem in autonomous driving as a sequence modeling task. The MTD-GPT model is trained and evaluated across several decision-making tasks, demonstrating performance that is either superior or comparable to that of state-of-the-art single-task decision-making models.

\end{abstract}

\section{INTRODUCTION}
Autonomous driving technology promises to bring revolutionary changes to transportation systems\cite{aradi2020survey,negash2022anticipation}, yet ensuring safe and accurate decision-making in complex scenarios remains a significant challenge for autonomous vehicles (AVs) \cite{chandra2022towards}. Intersections represent one of the most challenging driving scenarios, where decision-making typically encompasses tasks such as making left turns, proceeding straight, and making right turns. Extensive research has investigated decision-making and interaction issues for autonomous vehicles in intersection scenarios\cite{aradi2020survey}. In these studies, rule-based methods (such as PET\cite{de2017decision}) often struggle to handle decision-making in complex scenarios, and game theory-based methods typically require the setting of strong assumptions and pose challenges in terms of computational efficiency\cite{hang2020human,zhao2021yield}.

Reinforcement learning (RL), due to its exceptional learning capabilities and computational efficiency, has been widely applied in the design of decision-making algorithms for AV\cite{liu2022graph,kai2020multi,seong2021learning}. However, these methods mostly use a single model to handle different tasks, and utilizing a single model to cope with multiple autonomous driving decision-making scenarios and tasks remains a significant challenge for RL\cite{kai2020multi}.

To enhance the generalization capabilities and performance of RL, researchers have recently revisited and reformulated RL using Transformer models, achieving promising results\cite{chen2021decision}. 
Inspired by these works, we treat the autonomous driving multi-task decision-making problem as a sequence modeling and prediction problem, using the Generative Pre-trained Transformer (GPT) model to learn driving data and generate action decisions.

Concurrently, we introduce a pipeline for training GPT in multi-task decision-making, with guidance from RL experts: initially, we train an expert model for a single decision-making task using RL algorithms, followed by sampling expert data in the environment; ultimately, we utilize a mixed multi-task dataset for offline GPT training.
Based on the training pipeline and GPT-2\cite{radford2019language}, we propose a Multi-Task Decision-Making GPT (MTD-GPT) model for the multi-task decision-making of autonomous driving at unsignalized intersections, which is capable of simultaneously executing the decision-making tasks of turning left, going straight, and turning right at unsignalized intersections. 

We train multiple MTD-GPT models with different parameter scales and evaluate them in several decision-making tasks at intersections through simulations. We find that the performance of MTD-GPT across various tasks is superior to or on par with state-of-the-art (SOTA) single-task decision-making experts.

Our contributions can be summarized as follows:
\begin{itemize}
    \item We abstract the autonomous driving multi-task decision-making problem as a sequence modeling problem and propose a Multi-Task Decision-Making GPT (MTD-GPT) model based on GPT-2.
    \item We design a pipeline for training MTD-GPT, utilizing RL algorithms to train single-task decision-making experts and providing guidance for MTD-GPT learning using expert data.
    \item We assess MTD-GPT's performance on various decision-making tasks at unsignalized intersections, finding that our model's performance is either superior or comparable to that of outstanding single-task decision-making RL models.
\end{itemize}

\section{Related Works}

\subsection{Decision-making of AV at Intersection}

In recent years, considerable research has been devoted to the problems of decision-making and interaction of AV in intersection scenarios. These studies have employed various approaches, including ruled-based methods\cite{zhang2017finite}, game-theoretic methods\cite{cai2021game,hang2020human}, and data-driven techniques\cite{seong2021learning,kai2020multi,shu2021driving}, in which RL is recognized as a flexible, efficient, and potent method. However, its widespread implementation is hindered by several obstacles, one of which is training a RL model that can effectively manage a range of driving situations and decision-making tasks\cite{kai2020multi}. Kai et al. \cite{kai2020multi} formulated multitask objectives as a four-dimensional vector and devised a vectorized reward function to tackle multi-task decision-making issues at intersections. Liu et al. \cite{liu2022multi} introduced a multi-task safe reinforcement learning framework, resulting in more secure intersection decision-making for autonomous driving.

Nonetheless, these approaches are heavily dependent on the intricate design of state spaces and model components, posing difficulties when extending their application to broader and more complex scenarios. In this study, we utilize the GPT model for multi-task decision-making, which circumvents the need to create specific structures and components for individual subtasks, potentially improving the model's adaptability and generalization abilities.

\subsection{Transformer in Reinforcement Learning}
The remarkable accomplishments of Transformer models in domains such as natural language processing (NLP)\cite{casola2022pre} and computer vision (CV)\cite{khan2022transformers} in recent years have generated considerable interest.  Consequently, numerous researchers have attempted to apply Transformers to the RL domain, achieving remarkable results\cite{yuan2023transformer,zheng2022online,lin2022switch}. 
A notable approach entails converting RL problems into sequence modeling tasks and directly employing Transformer architectures for learning, which showcase exceptional performance and generalization properties\cite{chen2021decision,janner2021offline,lin2022switch}.

In this study, we recast the autonomous driving decision-making problem as a sequence modeling task and employ a Transformer architecture identical to that of GPT-2 to learn from the data, aiming to achieve multi-task decision-making at unsignalized intersections.

\section{Problem Formulation}
\subsection{Scenario Description}
\begin{figure}[!htbp]
    \centering
    \includegraphics[width=0.45\textwidth]{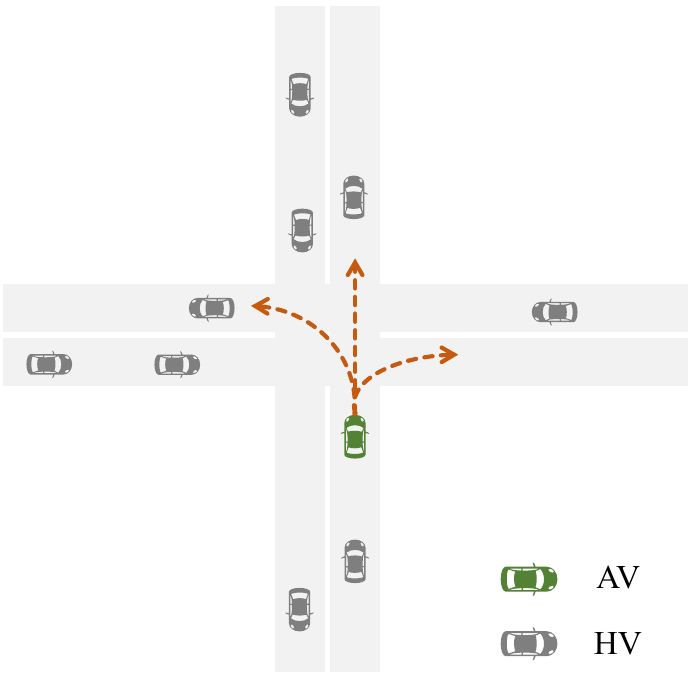}
    \caption{The intersection scenario for multi-task decision-making in our research.}
    \label{fig:multi_task_scenario}
\end{figure}

The problem of multi-task decision-making for AV at unsignalized intersections is considered. Specifically, we define a single-lane cross-shaped unsignalized intersection. A couple of human-driven vehicles (HVs) with different driving styles and intentions appear randomly from different directions and positions. After interacting with the HVs, the AV needs to complete the tasks of turning left, going straight, or turning right.
The scenario we study is shown in Fig.\ref{fig:multi_task_scenario}

\subsection{Vehicle Kinematics}
In our problem, AV actions determined by the RL expert and MTD-GPT model are converted to low-level steering and acceleration signals via a closed-loop PID controller.
Vehicle position and heading are controlled by Eq. \eqref{position control} and Eq. \eqref{heading control}, respectively:
\begin{equation}
    \label{position control}
    \begin{aligned}
        v_{\text{lat},r} &= -K_{p,\text{lat}} \Delta_{\text{lat}}, \\
        \Delta \psi_{r} &= \arcsin \left(\frac{v_{\text{lat},r}}{v}\right)
    \end{aligned}
\end{equation}
where
$\Delta_{\text{lat}}$ is the lateral position of the vehicle with respect to the lane center-line, $v_{\text{lat},r}$ is the lateral velocity command, $\Delta \psi_{r}$ is a heading variation to apply the lateral velocity command.
\begin{equation}
    \label{heading control}
    \begin{aligned}
        \psi_r &= \psi_L + \Delta \psi_{r}, \\
    \dot{\psi}_r &= K_{p,\psi} (\psi_r - \psi), \\
    \delta &= \arcsin \left(\frac{1}{2} \frac{l}{v} \dot{\psi}_r\right)
    \end{aligned}
\end{equation}
where $\psi_L$ is the lane heading, $\psi_r$ is the target heading to follow the lane heading and position,$\dot{\psi}_r$ is the yaw rate command,
$\delta$ is the front wheels angle control,$K_{p,\text{lat}}$ and $K_{p,\psi}$ are the position and heading control gains.

Vehicle motion is determined by a Kinematic Bicycle Model\cite{polack2017kinematic}:
\begin{equation}
\label{bicycle model}
\begin{aligned}
    \dot{x}&=v\cos(\psi+\beta) \\
    \dot{y}&=v\sin(\psi+\beta) \\
    \dot{v}&=a \\
    \dot{\psi}&=\frac{v}{l}\sin\beta \\
    \beta&=\tan^{-1}(1/2\tan\delta) \\
\end{aligned}
\end{equation}
where $(x, y)$ is the vehicle position, $v$ is forward speed, $\psi$ is heading, $a$ is the acceleration command, $\beta$ is the slip angle at the center of gravity, $\delta$ is the front wheel angle used as a steering command.

\subsection{Multi-Task Partially Observable Markov Decision Process}  
We describe the multi-task decision-making process of AV in the traffic environment as a multi-task partially observable Markov Decision Process (POMDP), which can be formulated by  $M = (\mathcal{S},\mathcal{A},\mathcal{O},\mathcal{P},\{r_i\}^N_{i=1})$, where $\mathcal{S}$ is the state space; $\mathcal{A}$ is the action space; $\mathcal{O}$ is the observation space; $\mathcal{P}$: $\mathcal{S} \times \mathcal{A} \rightarrow \mathbb{R}$ is the transition function; 
$\{r_i\}^N_{i=1}$ is a finite set of reward functions with different tasks,
$r_i$ denotes the reward function of task $i$. The goal of the offline GPT model is to find a policy $\pi_{gpt}(a|s)$ that maximizes expected return over all the tasks: 
\begin{equation}
    \pi^{*}_{gpt}(a|s)=\arg \max_{\pi}\mathbb{E}_{i\sim[N]}\mathbb{E}_{\pi} 
  \big [\sum_{t=1}^T{r_i(s_t,a_t)}  \big]  
\end{equation}

The observation space and action space of AV in our works are defined as follows:
\subsubsection{Observation Space}
The set of all observable HVs within the perception range of AV $i$ is denoted as $\mathcal{H}_i$. The observation matrix for expert $i$, denoted as $\mathcal{O}_i$, is a matrix of dimensions $| \mathcal{H}_i | \times | \mathcal{F} |$, where $|\mathcal{H}_i|$ represents the total number of observable vehicles for expert $i$, and $| \mathcal{F} |$ signifies the number of features used to describe a vehicle's state. The feature vector for vehicle $k$ is expressed as:
\begin{equation}
\mathcal{F}_k = [x_k, y_k, v^x_k, v^y_k]
\end{equation}
where $x_k, y_k, v^x_k, v^y_k$ correspond to the longitudinal position, lateral position, longitudinal speed, and lateral speed, respectively.
    
\subsubsection{Action Space}
We focus primarily on the high-level decision-making actions of AV. As the AV navigates through an intersection, we predefine the driving route, requiring the AV to determine acceleration and deceleration actions to reach its destination. Consequently, we define the action space $\mathcal{A}$ for the AV as a set of high-level control decisions, encompassing $\{ slow \ down, cruising, speed \ up \}$. Both acceleration and deceleration actions have absolute values of \SI{1}{m/s^2}.

\section{Methodology}
In this section, the training pipeline used to train MTD-GPT is first introduced, then the process of expert data collection is described. Finally, detailed information on our MTD-GPT model is introduced. 

\subsection{Overview of the Training Pipeline for MTD-GPT}
\begin{figure}[!htbp]
    \centering
    \includegraphics[width=0.45\textwidth]{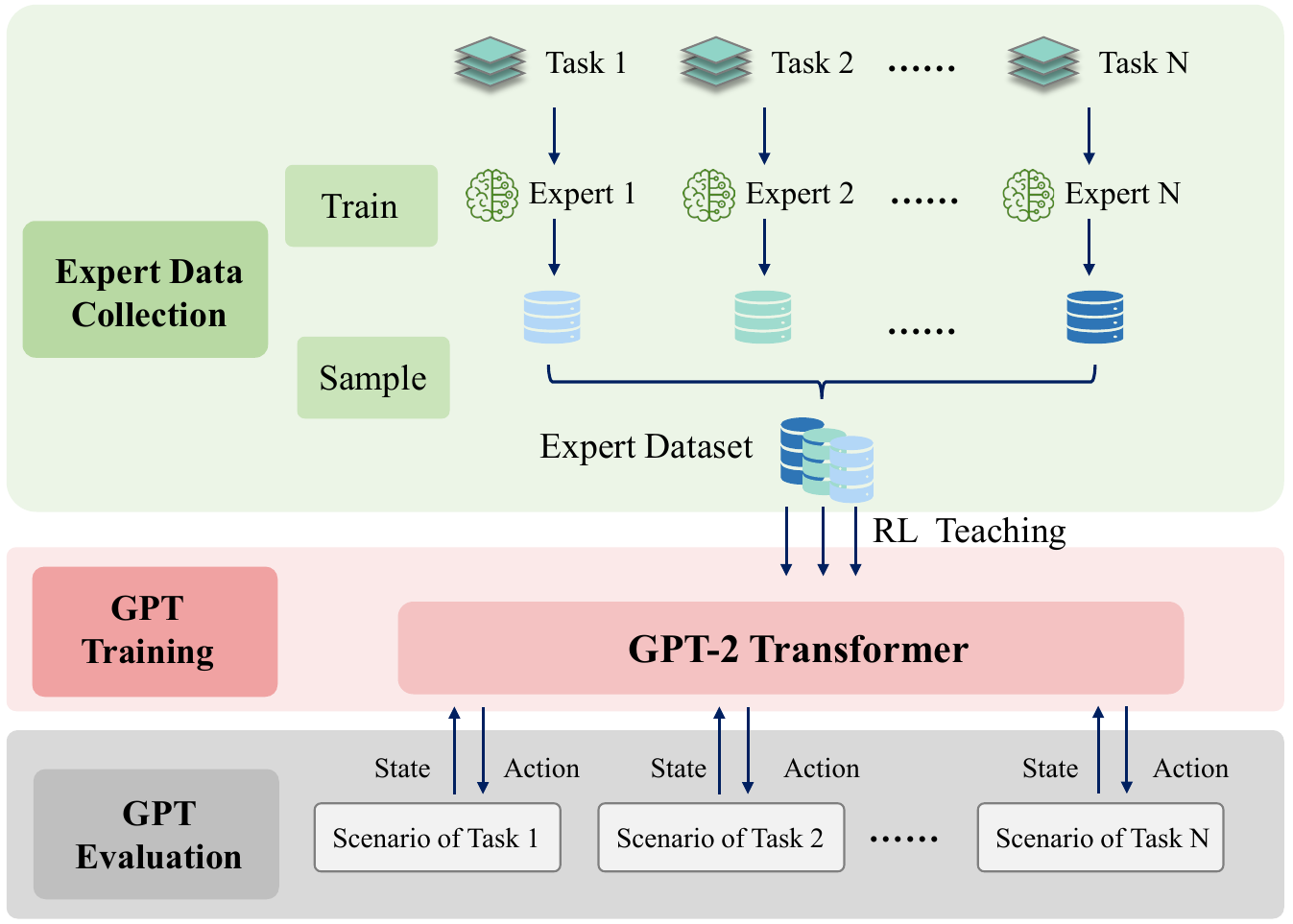}
    \caption{The Training Pipeline for our MTD-GPT.}
    \label{fig:training_pipeline}
\end{figure}

As shown in Fig. \ref{fig:training_pipeline}, MTD-GPT's training pipeline is composed of three key components: Expert Data Collection, GPT Training, and GPT Evaluation. During the Expert Data Collection phase, multiple expert models are trained using the proximal policy optimization (PPO) algorithm with the attention mechanism to achieve excellent performance on single-task decision-making. Subsequently, in a simulated environment, the expert's actions are recorded, generating a multi-task expert demonstration dataset.

In the GPT Training phase, the multi-task expert dataset serves as a guide for GPT learning. 
First, the problem of multi-task decision-making for autonomous driving is  abstracted as a sequence modeling and prediction task. Then we transform the "state-action-reward" tuples from expert data into a token format similar to Natural Language Processing (NLP) task to match the input format of the GPT model\cite{chen2021decision}.

In the GPT Evaluation phase, the trained MTD-GPT is assessed in various task scenarios, considering the decision-making data of GPT as quasi-expert data for future training of GPT. 

\subsection{Expert Data Collection}
\begin{figure}[!htbp]
    \centering
    \includegraphics[width=0.45\textwidth]{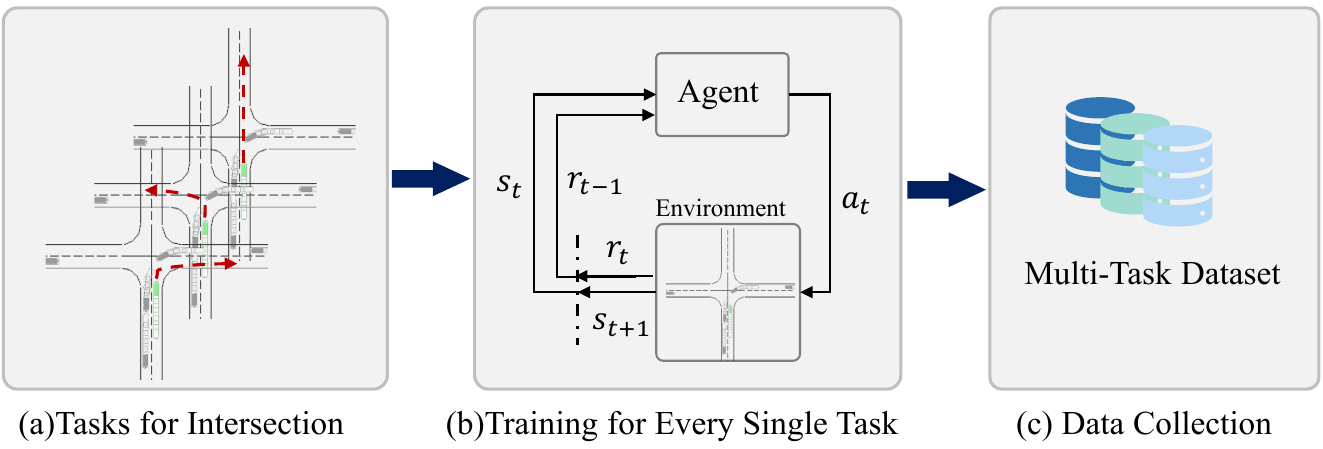}
    \caption{The process of the data sample for offline training.}
    \label{fig:data_sample}
\end{figure}

The entire process of expert data collection is illustrated in Fig. \ref{fig:data_sample}. Initially, three distinct autonomous vehicle decision-making tasks -- turning left, proceeding straight, and turning right -- are defined. Subsequently, three RL experts are trained using the PPO-Attention algorithm. Ultimately, the action and reward data are recorded and compiled into an offline multi-task dataset through simulating the actions of each expert within a designated simulation environment.

\subsubsection{RL Experts Training}
First, we train a couple of RL-based experts $i \ (i\in {1,2,...,N})$ who is good at single decision-making task.
Every RL expert consists of a policy network $\pi_{\theta}$ parameterized by $\theta$ and a value network $V_{\phi}$ parameterized by $\phi$. The policy network maps the observation $\mathcal{O}$ to a distribution of actions $\mathcal{a}$. And taking the same inputs as the policy network, the value network will estimate a scalar value $v$.

\begin{itemize}
    \item Policy Network.
    To better the performance of the RL Expert, the attention mechanism\cite{vaswani2017attention} is integrated into the policy network. The policy network with attention module is shown in Fig.\ref{fig:policy_network}. 
    For each RL expert $i$, the corresponding observation $\mathcal{O}_i$ and states $\mathcal{F}_i$ are initially embedded via a Multilayer Perceptron (MLP) encoder. Following this initial transformation, the embedded data undergoes further processing within an attention layer, serving to prioritize and capture salient features. Finally, the attention-focused data is decoded by an MLP decoder, converting the processed information into an actionable output.
    
    \begin{figure}[!htbp]
        \centering
        \includegraphics[width=0.45\textwidth]{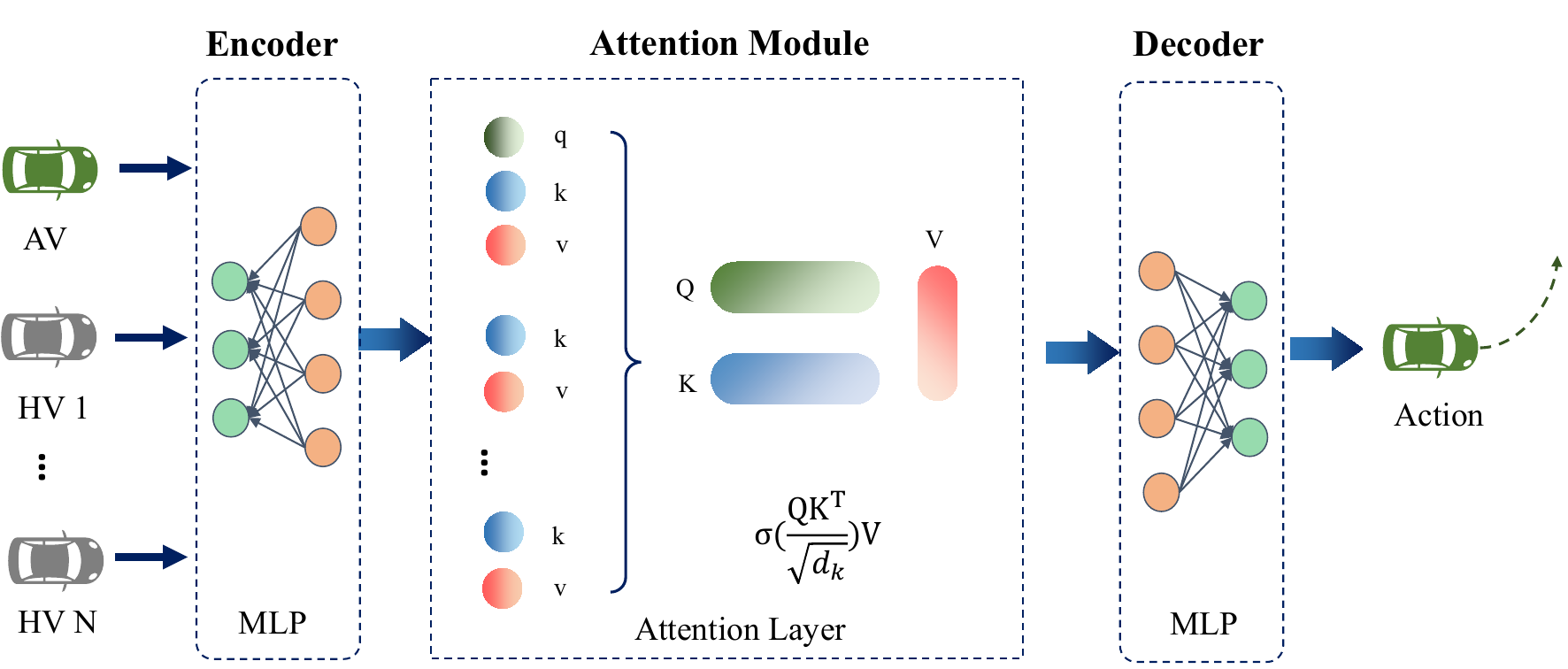}
        \caption{The policy network with attention layer.}
        \label{fig:policy_network}
    \end{figure}

    \item Policy Optimization. 
    The Clipped PPO\cite{schulman2017proximal} is used to train the policy network $\pi_{\theta}$ and the value network $V_{\phi}$, whose main idea is the clipping surrogate objective:
    \begin{equation}
    \begin{aligned}
        L^{\text{PPO}}(\theta)=E_t \Big[ \min \Big( r_t(\theta)\hat{A_t}, &\\
        \text{clip} \big (r_t(\theta),1-\epsilon,1+\epsilon \big) \hat{A_t} \Big) \Big]
    \end{aligned}
    \end{equation}
    where $r(\theta) = \frac{\pi_{\theta}(a|s)}{\pi_{\theta ^{\prime}(a|s)}}$ denotes the ratio of the new policy $\pi_{\theta}(a|s)$ to the old policy $\pi_{\theta ^{\prime}(a|s)}$,  $\hat{A_t}$ signifies the advantage function and $\epsilon$ is the clipping range.
    
    \item Reward Function. 
    To pass the intersection in a safe and efficient way, the reward function of expert $i$ is defined as:
    \begin{equation}
        r_{i} = w_c r_c + w_e r_e + w_a r_a
        \label{eq:reward}
    \end{equation}
    where $w_c$,$w_e$, and $w_a$ are the weight coefficients of collision reward $r_c$, efficiency reward $r_e$, and arrival reward $r_a$, respectively.
\end{itemize}

\subsubsection{Data Sample}
Upon completing the RL training, we deploy all expert models within the intersection simulation environment to collect a combined expert dataset $\mathcal{D} = \cup^N_{i=1} \mathcal{D}_i$. 
In each episode, the data sampling from experts is transformed into a sequential representation: $\tau = ( s_1, a_1,r_1, s_2, a_2,r_2,..., s_T, a_T,r_T )$, where $s_t$, $a_t$, $r_t$ are the state, action and reward of the expert at timestep t, respectively.

\begin{figure}[!htbp]
    \centering
    \includegraphics[width=0.45\textwidth]{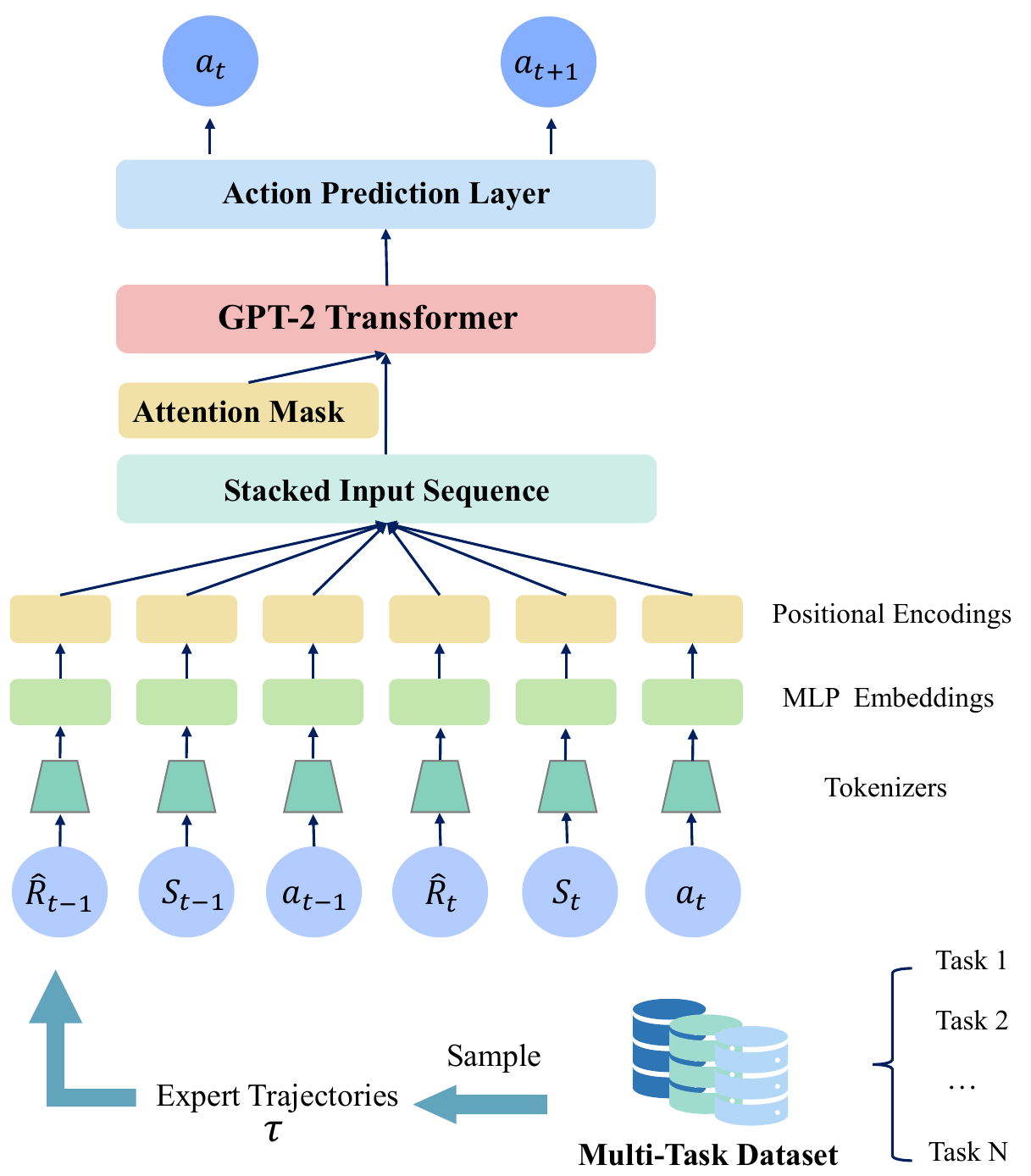}
    \caption{The GPT model streamlines AV decision-making by sampling expert trajectories, tokenizing them, mapping tokens via MLP and positional encoding, masking irrelevant information with self-attention, and predicting actions through a linear layer.}
    \label{fig:Model_GPT}
\end{figure}

\subsection{Offline Training GPT}
Following\cite{chen2021decision}, the GPT's training process is considered as the sequence modeling problem and will be trained in an autoregressive way. 

\subsubsection{Input Representation}
Let $\tau \ (\tau \in \mathcal{D})$ denote a trajectory of AV and let $|\tau|$ denote its length. The return-to-go (RTG) of the trajectory $\tau$ is defined as: $g_t = \sum^T_{t^{\prime} = t} r_{t^{\prime}} $, which represents the sum of future AV 's rewards from timestep $t$.
Let $s=(s_1,...,s_{|\tau|})$, $a = (a_1,...,a_{|\tau|})$ and $g = (g_1,...,g_{|\tau|})$ denote the sequence of  state, action and RTG of $\tau$, respectively.
Consequently, the representation of a trajectory sent to GPT is :
\begin{equation}
    \tau^{\prime} = \Big( s_1, a_1,g_1, s_2, a_2,g_2,..., s_T, a_T,g_T \Big)
\end{equation}
The initial RTG $g_1$ is equal to the return of the trajectory.

\subsubsection{Architecture}

As shown in Fig.\ref{fig:Model_GPT}, the MTD-GPT model utilizes an approach similar to Natural Language Processing (NLP) techniques for modeling and predicting decision-making tasks in autonomous driving. 

Initially, expert trajectories $\tau^{\prime}$ are randomly sampled from the data pool $\mathcal{D}$ for various subtasks, and these trajectories are then transformed into $tokens$ suitable for the model's processing, where $\langle s_t,a_t,g_t \rangle$ is defined as one $token \ x_t$ at timestep $t$.

Next, a MLP is utilized to map tokens to a continuous vector space and positional encodings (PE)\cite{vaswani2017attention} are then added to the embeddings to maintain the order of the input sequence:
\begin{equation}
    \mathcal{x}^{\prime}_t = \text{MLP}(x_t)
\end{equation}
\begin{equation}
    \mathcal{e}_t = \mathcal{x}^{\prime}_t + \text{Positional\_Encoding}(\mathcal{x}_t)
\end{equation}

Then the embedding results are passed through the Transformer's layers to acquire hidden states $h_t$:
\begin{equation}
    h_t = \text{TransformerLayer}(\mathcal{e}_t)
\end{equation}

\begin{figure}[!htbp]
    \centering
    \includegraphics[width=0.45\textwidth]{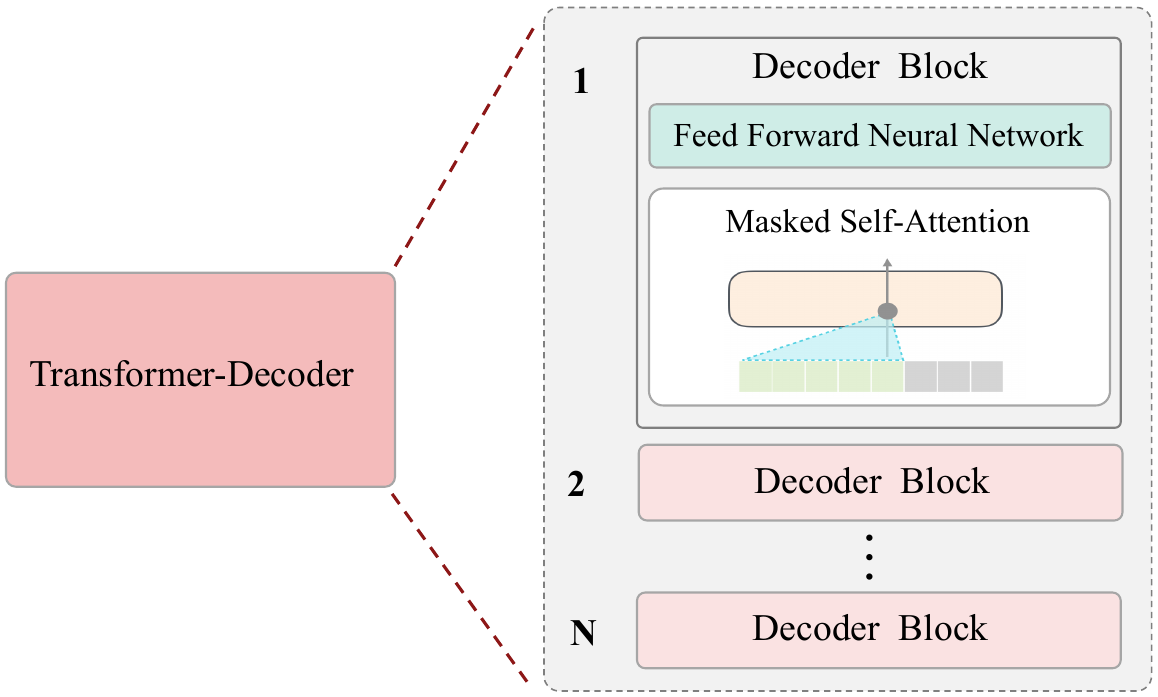}
    \caption{The original structure of GPT-2.}
    \label{fig:GPT_2_detail}
\end{figure}

As shown in Fig.\ref{fig:GPT_2_detail}, we use GPT-2 \cite {radford2019language} as our transformer model, which has the decoder-only architecture.
To mask irrelevant information, the self-attention mechanism\cite{vaswani2017attention} is applied, which computes the attention scores using $query \ Q$, $key \ K$,  and $value \ V$ matrices:
\begin{equation}
    Attention(Q,K,V) = softmax \Big ( \frac{QK^T}{\sqrt{d_k}} + M \Big) V
\end{equation}
where $M$ is the matrix that ensures the input trajectory at the timestep $t$ can only correlate with the input from $\langle 1, \cdots,t-1 \rangle$.

Finally, a linear prediction layer is used to generate predicted action $a_{t+1}$ based on the hidden states:
\begin{equation}
    a^{\prime}_{t+1} = \text{Linear}(h_t)
\end{equation}

By following this process, at timestep $t$, MTD-GPT generates the action  $a^{\prime}_{t+1}$ conditioned on the tokens from the latest $K$ timesteps, where $K$ is a hyperparameter and is also referred to as the $context \ length$ for the GPT.
Specifically, the policy of MTD-GPT is represented as $\pi_{gpt}(a_{t}|s_{-K,t},g_{-K,t})$, where $s_{-K,t}$ is shorthand for the sequence of $K$ past states $s_{\max (1,t-K+1):t}$ and similarly for $g_{-K,t}$. The whole training procedure of MTD-GPT is summarized in Algorithm \ref{algo:MTD_GPT_Train}

\subsubsection{Training}
All tokens from dataset $\mathcal{D}$ are fed into MTD-GPT, of which left-turn, straight-through, and right-turn tasks each account for one-third. The policy $\pi_{gpt}$ is trained by the Cross-entropy (CE) loss:
\begin{equation}
\label{eq:Loss_function}
    \mathcal{L}_{CE} = \frac{1}{K} \sum^K_{t=1} P(a_t)\log \Big( \pi_{gpt}(s_{-K,t},g_{-K,t})  \Big )
\end{equation}

\subsection{GPT Evaluation}
After training, MTD-GPT will be evaluated in different driving tasks at intersections.
For task $i$, we specify the desired performance $g^i_1$ and an initial state $s^i_1$ of AV. The MTD-GPT generates the action $a^i_1 = \pi_{gpt}(s^i_1,g^i_1)$. Then the action $a^i_1$ will be executed by AV. 
The next state $s^i_{t+1} \sim  P(\cdot |s^i_t,a^i_t)$ and a reward $r^i_t = R(s^i_t,a^i_t)$ will be obtained, which gives us the next RTG as $g^i_{t+1} = g^i_t - r^i_t$. And GPT generates the action $a^i_2$ based on $s^i_1, s^i_2$ and $g^i_1,g^i_2$. The whole decision-making process is repeated until the episode terminates.
Then the decision success rate of the AV on each task will be tallied and calculated.

\begin{algorithm}
\SetAlFnt{\small}
    \SetKwInOut{Parameter}{Inputs}
    \SetKwInOut{Output}{Outputs}
\SetKwRepeat{Repeat}{repeat}{until}
\caption{Multi-Task Decision-Making GPT}
\label{algo:MTD_GPT_Train}
\LinesNumbered 
\SetAlgoLined
\Parameter{Offline Dataset $\mathcal{D}$}
\Output{ $\pi_{gpt}(a|s,g)$}
\vspace{0.2em}
\hrule
\vspace{0.2em}
Initialize GPT model with random weights $\theta$;\\
\For{$Epoch = 1$ to $M$}
{
\Repeat{$\mathcal{D}$ is empty}
{
    Sample a batch of trajectories $\tau$ of task $i$ from $\mathcal{D}$;\\
    Compute RTG $g$ for each trajectory $\tau$;\\
    Get $tokens \ x$ : $Tokenize(\tau)$;\\
    Embed input tokens : $e = MLP(x) + PE_t$;\\
    Acquire hidden states $h$ by $\text{TransformerLayer}(e)$;\\
    Get predicted action $a$ by $\text{LinearLayer}(h)$;\\
    Compute CE loss $L_{CE}(\theta)$;\\
    Update $\theta$ using gradient descent on $L_{CE}(\theta)$;\\
}
}
\end{algorithm}

\section{Simulation and Performance Evaluation}
\subsection{Simulation Environment}
Our simulation platform is built based on an OpenAI Gym environment\cite{highway-env}. The longitude and lateral decisions of HVs are controlled by the IDM\cite{kesting2010enhanced} and MOBIL\cite{kesting2007general} models, respectively. 
All HVs in our simulator are set with the constant-speed motion prediction and collision avoidance functions.
\begin{figure*}[!htbp]
    \centering
    \includegraphics[width=0.95\textwidth]{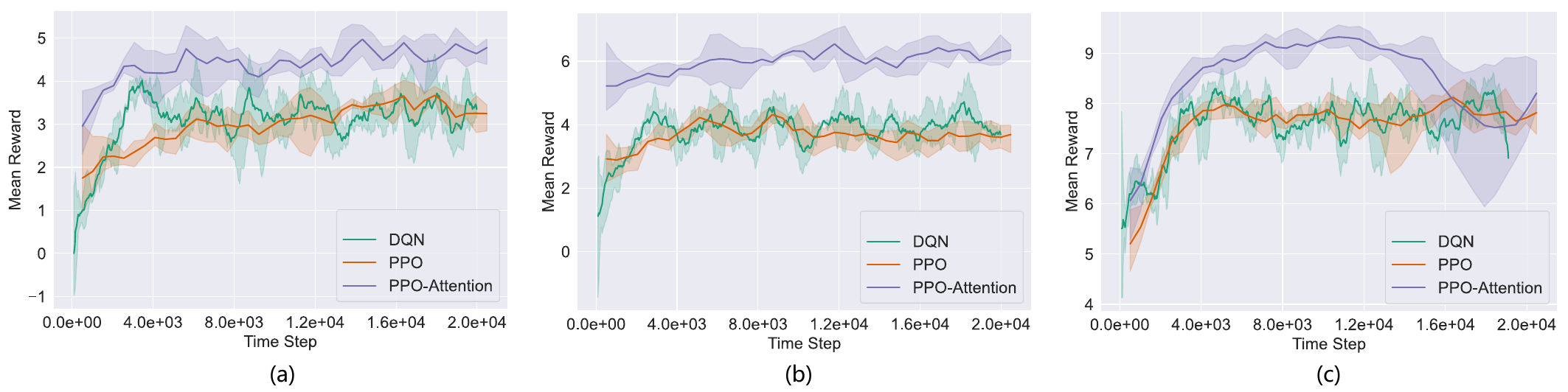}
    \caption{The performance comparison between RL Experts (PPO-Attention) and other baselines:(a) turning left task, (b) going straight task, (c) turning right task.}
    \label{fig:RL_compare}
\end{figure*}

\begin{figure*}[!htbp]
    \centering
    \includegraphics[width=0.95\textwidth]{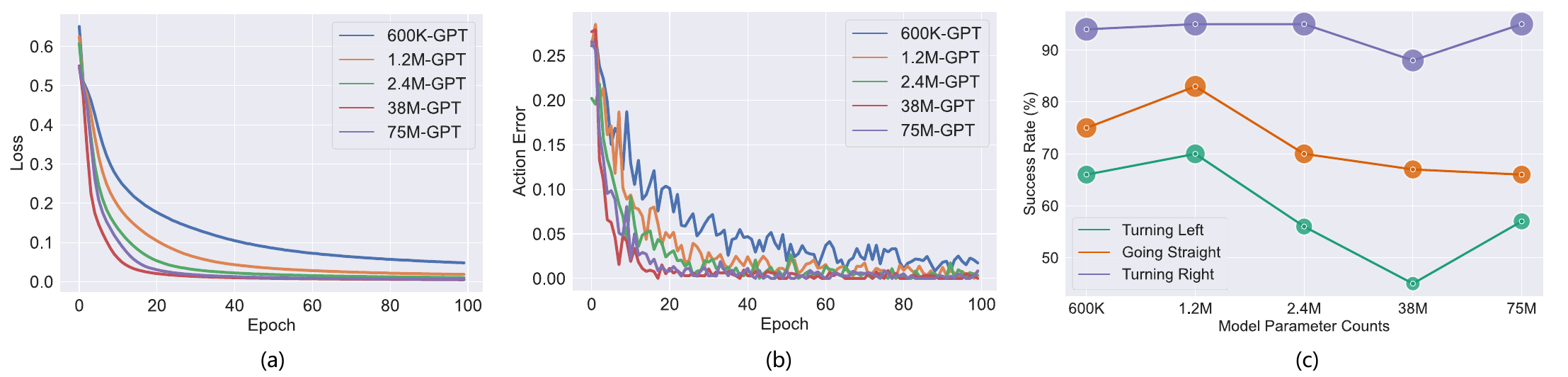}
    \caption{The performance of GPT with different parameter counts: (a) training loss, (b) training action error,  (c) the success rate on different testing tasks.}
    \label{fig:compare_diff_parameter}
\end{figure*}

\subsection{Implementation Details}
For the RL expert, the Encoder and Decoder of the attention-based policy network both are MLP, which has two linear layers and the layer size is $64 \times 64$. The Attention Layer contains $2$ attention heads and the feature size is $128$. We use Deep Q-learning (DQN) and Proximal Policy Optimization (PPO) as our single-task baselines. The total training steps are all $20K$. The $w_c$, $w_e$ and $w_a$ in Eq.\eqref{eq:reward} are all set as 1.
The parameters of MTD-GPT are shown in Table \ref{tab:training_hyperparameter}.

All experiments are conducted in a computation platform with Intel Xeon Silver 4214R CPU, NVIDIA GeForce RTX 3090 GPU$\times$2, and 128G Memory.
\begin{table}[!htbp]
    \centering
    \caption{The hyperparameter of the MTD-GPT model}
    \label{tab:training_hyperparameter}
    \begin{tabular}{c c c}
        \toprule
        Symbol & Definition & Value\\
        \midrule
        $M$ & Training Epoch & 100 \\
        $N_s$ & Steps for Each Epoch & $10^4$ \\
        $K$  & Training Context Length & 30 \\
        $D$  & Dropout   & 0.1 \\
        $N_h$ & Number of Attention Heads & 4 \\
        $N_l$  & Number of Layers  &  3/6/12 \\
        $E_d$  & Embedding Dimension  &  128/ 256/ 1024 \\
        $B_s$  & Batch Size & 64 \\
        \bottomrule
    \end{tabular}
\end{table}

\subsection{Performance Evaluation}
\subsubsection{RL Expert}
Fig. \ref{fig:RL_compare} presents a comparative analysis of the RL Expert algorithm, PPO-Attention, which we develop, juxtaposed with several baseline algorithms including DQN and PPO. Assessing both the speed of convergence and the average reward, it becomes discernible that PPO-Attention exhibits superior overall performance across all decision-making tasks. This indicates the robustness of our RL expert when faced with singular decision-making tasks, thereby supplying high-quality action data for GPT model.

\subsubsection{MTD-GPT Performance Analysis}
We train different models with varying parameter counts (approx. $600K, 1.2M, 2.4M, 38M, 75M$) by adjusting the number of decoder layers and embedding dimensions. The  training losses, action errors, and success rates on test tasks during the training process for GPT models with different parameter scales are depicted in Fig.\ref{fig:compare_diff_parameter} (a), Fig.\ref{fig:compare_diff_parameter} (b), and Fig.\ref{fig:compare_diff_parameter}(c), respectively.

As the model's parameter count increases, convergence is achieved faster during the training process. The two models with $38M$ and $75M$ parameters exhibit the best convergence and learning outcomes during training. 
However, we find that larger models do not necessarily guarantee better decision-making performance. We test these models on three decision tasks at intersections. 
MTD-GPT (600K) achieves decision success rates of $66\%$, $75\%$, and $94\%$ for left-turn, straight-ahead, and right-turn tasks, respectively. When the model's parameter count increases to 1.2M, the decision success rates increase by $4\%$, $8\%$, and $1\%$, respectively. Nevertheless, when the model's parameter scale further increases, the decision success rates actually decline.

We posit that models with larger parameter counts are more prone to overfitting on fixed offline data, leading to suboptimal performance in test tasks. Therefore, devising training strategies to prevent overfitting and exploring other methods to enhance model generalization capabilities will be our next area of focus.

\begin{figure}[!htbp]
    \centering
    \includegraphics[width=0.45\textwidth]{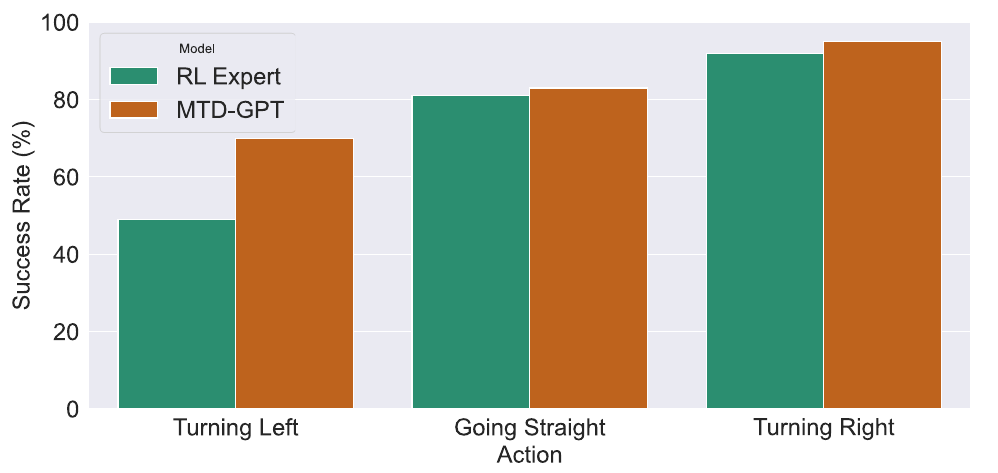}
    \caption{The performance comparison between GPT and RL Experts.}
    \label{fig:Performance_Compare}
\end{figure}

\begin{figure*}[!htbp]
    \centering
    \includegraphics[width=0.95 \textwidth]{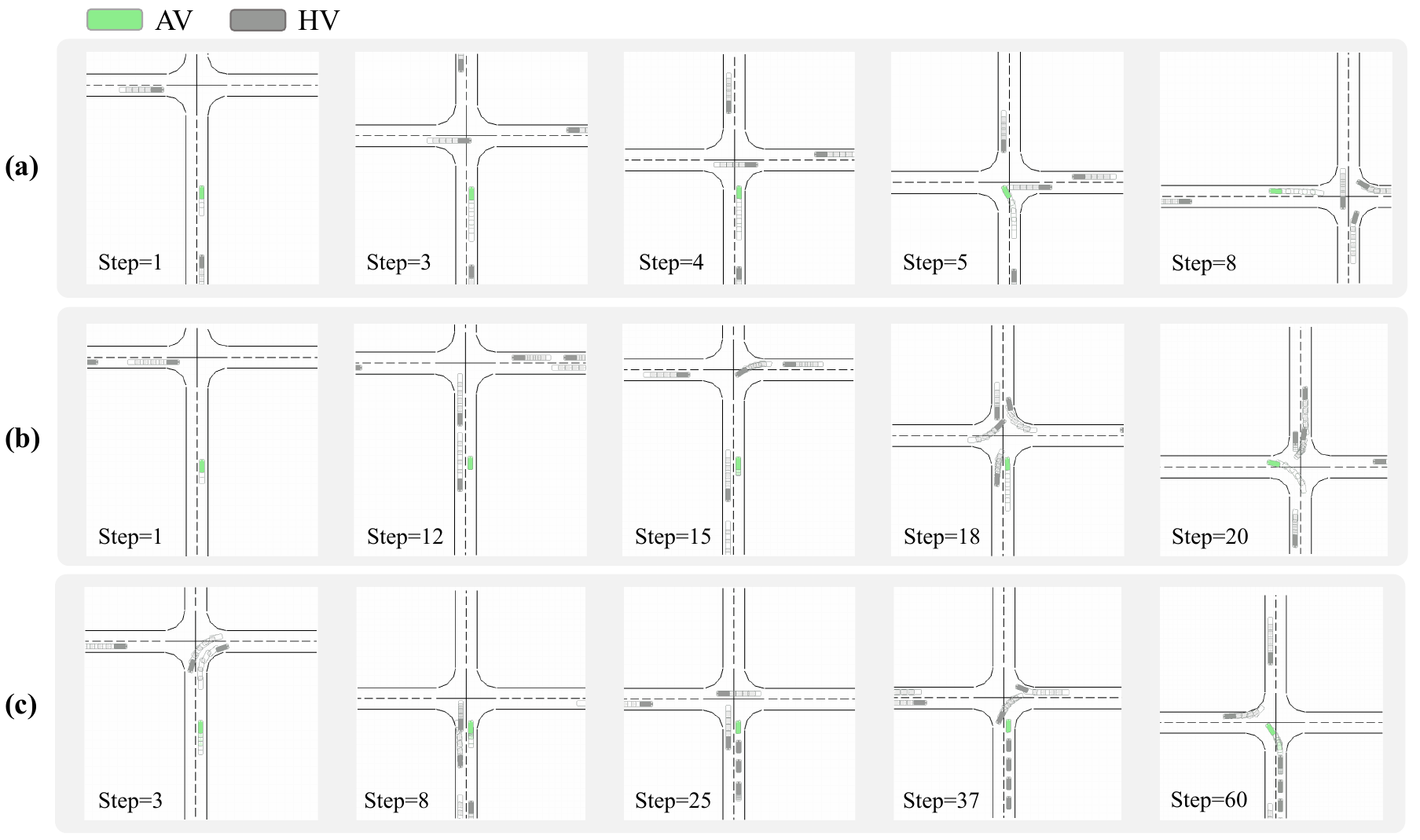}
    \caption{Three cases of turning left task from different models: (a) Case 1 from GPT with $600K$ parameters, (b) Case 2 from GPT with $1.2M$ parameters, (c) Case 3 from GPT with $75M$ parameters.}
    \label{fig:case_analysis}
\end{figure*}

\subsubsection{MTD-GPT vs. RL Expert}
As illustrated in Fig.\ref{fig:Performance_Compare}, we compare the single-subtask success rates of MTD-GPT (1.2M) and RL Expert. RL Expert achieves decision success rates of $49\%$, $81\%$, and $92\%$ for left-turn, right-turn, and straight-ahead tasks, respectively, while MTD-GPT (1.2M) achieves $70\%$, $83\%$, and $95\%$. These results show improvement compared to the single-task RL Expert, and even GPT models with other parameter scales can achieve performance comparable to RL Expert in these subtasks. This demonstrates the promising impact and significant potential of the MTD-GPT model in multi-task decision-making.

\subsection{Case Analysis}
Given the complexity involved in executing left turns at unsignalized intersections, we evaluate the performance of our model in such scenarios. We conduct an analysis on three distinct cases, each corresponding to a GPT model of different parameter scales ($600K, 1.2M, and 75M$). Animations demonstrating these cases and more demos can be accessed at the site.\footnote{See \url{https://shorturl.at/eMNS4}}

Our analysis reveals that, even when confronted with identical decision-making tasks, models with varying parameter scales demonstrate divergent decision-making styles. As depicted in Fig. \ref{fig:case_analysis}(a), for Case 1, the AV hardly reduces its speed after entering the intersection ($Step=3$) and navigates its way through just before an oncoming vehicle traveling straight arrives at the intersection ($Step=5$). This approach enables the AV to accomplish the task in less time, but the aggressive decision-making style heightens safety risks, indicating that it is not an optimal strategy.

Conversely, in Case 2, the AV adopts a more conservative behavior, as illustrated in Fig. \ref{fig:case_analysis}(b). Upon entering the intersection, the AV opts to halt and observe ($Step=12$), only advancing into the intersection when the risk is deemed low or when there is an absence of interaction conflict (i.e., after the oncoming vehicle has passed) ($Step=18$), thereby ensuring a safe departure from the intersection.

Furthermore, in the model with $75M$ parameters, we notice a further amplification of this conservative decision-making inclination. As demonstrated in Fig. \ref{fig:case_analysis}(c) for Case 3, the AV pauses before entering the intersection ($Step=8$). However, the adoption of an overly cautious strategy culminates in an extended waiting period ($Step=8-37$), subsequently leading to congestion in its lane and a decrease in the overall traffic system efficiency.

\section{CONCLUSIONS}
Developing a multi-task decision-making model for autonomous driving constitutes a remarkably challenging goal in the field of research. In this work, we propose the MTD-GPT model for multi-task decision-making of AV at unsignalized intersections.  
Moreover, we design a pipeline that leverages RL algorithms to train single-task decision-making experts and utilize expert data to provide guidance for MTD-GPT learning. Our experimental results demonstrate that the MTD-GPT model outperforms or matches the performance of exceptional single-task decision-making RL models in different decision-making tasks.

In future work, we aim to further expand the generalization capabilities of the MTD-GPT model, enabling exceptional performance in tasks such as merging onto ramps, navigating roundabouts, and changing lanes or overtaking on urban roads. Additionally, we plan to employ a hybrid dataset comprising natural driving and simulation data for GPT training. Furthermore, we will investigate incorporating Reinforcement Learning with Human Feedback (RLHF) methods into the GPT model's training regimen, promoting the generation of more human-like, secure, and interpretable decision-making behaviors.


\bibliographystyle{IEEEtran}  
\bibliography{reference}

\begin{thebibliography}{10}
\providecommand{\url}[1]{#1}
\csname url@samestyle\endcsname
\providecommand{\newblock}{\relax}
\providecommand{\bibinfo}[2]{#2}
\providecommand{\BIBentrySTDinterwordspacing}{\spaceskip=0pt\relax}
\providecommand{\BIBentryALTinterwordstretchfactor}{4}
\providecommand{\BIBentryALTinterwordspacing}{\spaceskip=\fontdimen2\font plus
\BIBentryALTinterwordstretchfactor\fontdimen3\font minus
  \fontdimen4\font\relax}
\providecommand{\BIBforeignlanguage}[2]{{%
\expandafter\ifx\csname l@#1\endcsname\relax
\typeout{** WARNING: IEEEtran.bst: No hyphenation pattern has been}%
\typeout{** loaded for the language `#1'. Using the pattern for}%
\typeout{** the default language instead.}%
\else
\language=\csname l@#1\endcsname
\fi
#2}}
\providecommand{\BIBdecl}{\relax}
\BIBdecl

\bibitem{aradi2020survey}
S.~Aradi, ``Survey of deep reinforcement learning for motion planning of
  autonomous vehicles,'' \emph{IEEE Transactions on Intelligent Transportation
  Systems}, vol.~23, no.~2, pp. 740--759, 2020.

\bibitem{negash2022anticipation}
N.~M. Negash and J.~Yang, ``Anticipation-based autonomous platoon control
  strategy with minimum parameter learning adaptive radial basis function
  neural network sliding mode control,'' \emph{SAE International Journal of
  Vehicle Dynamics, Stability, and NVH}, vol.~6, no. 10-06-03-0017, pp.
  247--265, 2022.

\bibitem{chandra2022towards}
R.~Chandra, ``Towards autonomous driving in dense, heterogeneous, and
  unstructured traffic,'' Ph.D. dissertation, University of Maryland, College
  Park, 2022.

\bibitem{de2017decision}
P.~De~Beaucorps, T.~Streubel, A.~Verroust-Blondet, F.~Nashashibi, B.~Bradai,
  and P.~Resende, ``Decision-making for automated vehicles at intersections
  adapting human-like behavior,'' in \emph{2017 IEEE Intelligent Vehicles
  Symposium (IV)}.\hskip 1em plus 0.5em minus 0.4em\relax IEEE, 2017, pp.
  212--217.

\bibitem{hang2020human}
P.~Hang, C.~Lv, Y.~Xing, C.~Huang, and Z.~Hu, ``Human-like decision making for
  autonomous driving: A noncooperative game theoretic approach,'' \emph{IEEE
  Transactions on Intelligent Transportation Systems}, vol.~22, no.~4, pp.
  2076--2087, 2020.

\bibitem{zhao2021yield}
X.~Zhao, Y.~Tian, and J.~Sun, ``Yield or rush? social-preference-aware driving
  interaction modeling using game-theoretic framework,'' in \emph{2021 IEEE
  International Intelligent Transportation Systems Conference (ITSC)}.\hskip
  1em plus 0.5em minus 0.4em\relax IEEE, 2021, pp. 453--459.

\bibitem{liu2022graph}
Q.~Liu, X.~Li, Z.~Li, J.~Wu, G.~Du, X.~Gao, F.~Yang, and S.~Yuan, ``Graph
  reinforcement learning application to co-operative decision-making in mixed
  autonomy traffic: Framework, survey, and challenges,'' \emph{arXiv preprint
  arXiv:2211.03005}, 2022.

\bibitem{kai2020multi}
S.~Kai, B.~Wang, D.~Chen, J.~Hao, H.~Zhang, and W.~Liu, ``A multi-task
  reinforcement learning approach for navigating unsignalized intersections,''
  in \emph{2020 IEEE Intelligent Vehicles Symposium (IV)}.\hskip 1em plus 0.5em
  minus 0.4em\relax IEEE, 2020, pp. 1583--1588.

\bibitem{seong2021learning}
H.~Seong, C.~Jung, S.~Lee, and D.~H. Shim, ``Learning to drive at unsignalized
  intersections using attention-based deep reinforcement learning,'' in
  \emph{2021 IEEE International Intelligent Transportation Systems Conference
  (ITSC)}.\hskip 1em plus 0.5em minus 0.4em\relax IEEE, 2021, pp. 559--566.

\bibitem{chen2021decision}
L.~Chen, K.~Lu, A.~Rajeswaran, K.~Lee, A.~Grover, M.~Laskin, P.~Abbeel,
  A.~Srinivas, and I.~Mordatch, ``Decision transformer: Reinforcement learning
  via sequence modeling,'' \emph{Advances in neural information processing
  systems}, vol.~34, pp. 15\,084--15\,097, 2021.

\bibitem{radford2019language}
A.~Radford, J.~Wu, R.~Child, D.~Luan, D.~Amodei, I.~Sutskever \emph{et~al.},
  ``Language models are unsupervised multitask learners,'' \emph{OpenAI blog},
  vol.~1, no.~8, p.~9, 2019.

\bibitem{zhang2017finite}
M.~Zhang, N.~Li, A.~Girard, and I.~Kolmanovsky, ``A finite state machine based
  automated driving controller and its stochastic optimization,'' in
  \emph{Dynamic Systems and Control Conference}, vol. 58288.\hskip 1em plus
  0.5em minus 0.4em\relax American Society of Mechanical Engineers, 2017, p.
  V002T07A002.

\bibitem{cai2021game}
J.~Cai, P.~Hang, and C.~Lv, ``Game theoretic modeling and decision making for
  connected vehicle interactions at urban intersections,'' in \emph{2021 6th
  IEEE International Conference on Advanced Robotics and Mechatronics
  (ICARM)}.\hskip 1em plus 0.5em minus 0.4em\relax IEEE, 2021, pp. 874--880.

\bibitem{shu2021driving}
H.~Shu, T.~Liu, X.~Mu, and D.~Cao, ``Driving tasks transfer using deep
  reinforcement learning for decision-making of autonomous vehicles in
  unsignalized intersection,'' \emph{IEEE Transactions on Vehicular
  Technology}, vol.~71, no.~1, pp. 41--52, 2021.

\bibitem{liu2022multi}
Y.~Liu, Y.~Gao, Q.~Zhang, D.~Ding, and D.~Zhao, ``Multi-task safe reinforcement
  learning for navigating intersections in dense traffic,'' \emph{Journal of
  the Franklin Institute}, 2022.

\bibitem{casola2022pre}
S.~Casola, I.~Lauriola, and A.~Lavelli, ``Pre-trained transformers: An
  empirical comparison,'' \emph{Machine Learning with Applications}, vol.~9, p.
  100334, 2022.

\bibitem{khan2022transformers}
S.~Khan, M.~Naseer, M.~Hayat, S.~W. Zamir, F.~S. Khan, and M.~Shah,
  ``Transformers in vision: A survey,'' \emph{ACM computing surveys (CSUR)},
  vol.~54, no. 10s, pp. 1--41, 2022.

\bibitem{yuan2023transformer}
W.~Yuan, J.~Chen, S.~Chen, L.~Lu, Z.~Hu, P.~Li, D.~Feng, F.~Liu, and J.~Chen,
  ``Transformer in reinforcement learning for decision-making: A survey,''
  2023.

\bibitem{zheng2022online}
Q.~Zheng, A.~Zhang, and A.~Grover, ``Online decision transformer,'' in
  \emph{International Conference on Machine Learning}.\hskip 1em plus 0.5em
  minus 0.4em\relax PMLR, 2022, pp. 27\,042--27\,059.

\bibitem{lin2022switch}
Q.~Lin, H.~Liu, and B.~Sengupta, ``Switch trajectory transformer with
  distributional value approximation for multi-task reinforcement learning,''
  \emph{arXiv preprint arXiv:2203.07413}, 2022.

\bibitem{janner2021offline}
M.~Janner, Q.~Li, and S.~Levine, ``Offline reinforcement learning as one big
  sequence modeling problem,'' \emph{Advances in neural information processing
  systems}, vol.~34, pp. 1273--1286, 2021.

\bibitem{polack2017kinematic}
P.~Polack, F.~Altch{\'e}, B.~d'Andr{\'e}a Novel, and A.~de~La~Fortelle, ``The
  kinematic bicycle model: A consistent model for planning feasible
  trajectories for autonomous vehicles?'' in \emph{2017 IEEE intelligent
  vehicles symposium (IV)}.\hskip 1em plus 0.5em minus 0.4em\relax IEEE, 2017,
  pp. 812--818.

\bibitem{vaswani2017attention}
A.~Vaswani, N.~Shazeer, N.~Parmar, J.~Uszkoreit, L.~Jones, A.~N. Gomez,
  {\L}.~Kaiser, and I.~Polosukhin, ``Attention is all you need,''
  \emph{Advances in neural information processing systems}, vol.~30, 2017.

\bibitem{schulman2017proximal}
J.~Schulman, F.~Wolski, P.~Dhariwal, A.~Radford, and O.~Klimov, ``Proximal
  policy optimization algorithms,'' \emph{arXiv preprint arXiv:1707.06347},
  2017.

\bibitem{highway-env}
E.~Leurent, ``An environment for autonomous driving decision-making,''
  \url{https://github.com/eleurent/highway-env}, 2018.

\bibitem{kesting2010enhanced}
A.~Kesting, M.~Treiber, and D.~Helbing, ``Enhanced intelligent driver model to
  access the impact of driving strategies on traffic capacity,''
  \emph{Philosophical Transactions of the Royal Society A: Mathematical,
  Physical and Engineering Sciences}, vol. 368, no. 1928, pp. 4585--4605, 2010.

\bibitem{kesting2007general}
------, ``General lane-changing model mobil for car-following models,''
  \emph{Transportation Research Record}, vol. 1999, no.~1, pp. 86--94, 2007.

\end{thebibliography}

\end{document}